\documentclass[runningheads]{llncs}
\usepackage[dvips]{graphicx}
\usepackage{url}
\usepackage{subcaption} 
\usepackage[table]{xcolor} 
\usepackage{amsmath}
\usepackage{amssymb}
\usepackage{makecell}       
\usepackage{adjustbox}
\usepackage{graphicx}     
\usepackage{multirow}      
\usepackage{booktabs}       
\definecolor{luminousGreen}{RGB}{57,255,20} 
\definecolor{luminousRed}{RGB}{255,0,0}      
\definecolor{pink}{RGB}{255,105,180}        
\usepackage[colorlinks=true,        
            citecolor=luminousGreen,  
            linkcolor=luminousRed,           
            urlcolor=pink]{hyperref}
\usepackage{marvosym}

\usepackage{subcaption}
\begin{document}
\title{DSGC-Net: A Dual-Stream Graph Convolutional Network for Crowd Counting via Feature Correlation Mining}
\titlerunning{DSGC-Net for Crowd Counting}
%
\author{Yihong Wu\inst{1} \and
Jinqiao Wei\inst{1} \and
Xionghui Zhao\inst{1} \and
Yidi Li\inst{1}\textsuperscript{(\Letter)} \and
Shaoyi Du\inst{2}\and
Bin Ren\inst{3,4} \and
Nicu Sebe\inst{3}}

\authorrunning{Y. Wu et al.}
%
\institute{Taiyuan University of Technology, Taiyuan 030024, China\\
\email{liyidi@tyut.edu.cn}
\and
Department of Ultrasound, the Second Affiliated Hospital of Xi’an Jiaotong University, State Key Laboratory of Human-Machine Hybrid Augmented Intelligence, and Institute of Artificial Intelligence and Robotics, Xi'an Jiaotong University, Xi'an 710049, China
\and
University of Trento, Trento, Italy
\and
University of Pisa, Pisa, Italy}


%
\maketitle              

\begin{abstract}
Deep learning-based crowd counting methods have achieved remarkable progress in recent years. However, in complex crowd scenarios, existing models still face challenges when adapting to significant density distribution differences between regions. Additionally, the inconsistency of individual representations caused by viewpoint changes and body posture differences further limits the counting accuracy of the models. To address these challenges, we propose DSGC-Net, a Dual-Stream Graph Convolutional Network based on feature correlation mining. 
DSGC-Net introduces a Density Approximation (DA) branch and a Representation Approximation (RA) branch. 
By modeling two semantic graphs, 
it captures the potential feature correlations in density variations and representation distributions.
The DA branch incorporates a density prediction module that generates the density distribution map, and constructs a density-driven semantic graph based on density similarity. The RA branch establishes a representation-driven semantic graph by computing global representation similarity. Then, graph convolutional networks are applied to the two semantic graphs separately to model the latent semantic relationships, which enhance the model's ability to adapt to density variations and improve counting accuracy in multi-view and multi-pose scenarios. Extensive experiments on three widely used datasets demonstrate that DSGC-Net outperforms current state-of-the-art methods. In particular, we achieve MAE of 48.9 and 5.9 in ShanghaiTech Part A and Part B datasets, respectively.
The released code is available at: \url{https://github.com/Wu-eon/CrowdCounting-DSGCNet}.

\end{abstract}
\section{Introduction}

Crowd counting aims to estimate the number of individuals in a given image \cite{sindagi2018survey}, with applications spanning multiple domains including security surveillance, event management, and traffic flow analysis \cite{khan2020advances,cardone2014crowdsensing,Li_Liu_Tang_2022}. 
Benefiting from the rapid advancement of deep learning, crowd counting technology has achieved significant improvements in accuracy \cite{wang2025comprehensive,cao2018scale}.  
Current mainstream methods can be categorized into two classes: density map-based methods \cite{pham2015count,li2018csrnet} and point regression-based methods \cite{song2021rethinking,liang2022end,liu2024consistency}. 
Density map-based methods estimate crowd counts through density distribution integration, excelling in high-density scenarios, while point regression approaches directly localize individuals using annotated points, proving more effective for sparse scenes.  

Despite the remarkable progress achieved in recent years, crowd counting in complex real-world settings faces significant challenges due to density variations, perspective differences, and posture diversity \cite{savner2023crowdformer,alazbah2023exploring}.  
On the one hand, uneven crowd density distributions challenge counting accuracy in complex scenes \cite{gao2020cnn}. Specifically, in sparse areas, individuals are often overlooked, while dense regions suffer from missed and duplicate counts due to occlusion and blurring \cite{grant2017crowd}. On the other hand, the ``near-large, far-small'' perspective effect creates substantial representation differences between near and distant individuals, compounded by diverse person postures, further complicating feature learning.
Existing models \cite{song2021rethinking,abousamra2021localization,wang2020distribution} often overfit to specific representation types, limiting their adaptability and generalization across multi-view and multi-posture scenarios. Although multi-scale feature fusion \cite{sindagi2019multi,10814658} improves adaptability to scale and perspective variations, partially mitigating the aforementioned issues, there remains considerable room for improvement in counting accuracy.

To address the above challenges, Graph Convolutional Networks (GCNs) \cite{kipf2016semi} have emerged as a highly promising solution. Graph structures flexibly model topological relationships in the spatial domain, while GCNs, through iterative feature transformations and information propagation, capture deep feature correlations on a global scale. Although recent studies have begun applying graph modeling approaches to visual recognition tasks, their potential for mining such correlations remains far from fully explored \cite{miao2024multi}.

In this paper, a Dual-Stream Graph Convolutional Network (DSGC-Net) is proposed to address the challenges posed by density variations and individual representation diversity.
DSGC-Net is a hybrid architecture that integrates both density map and point regression strategies. 
It consists of two parallel branches: the Density Approximation (DA) branch and the Representation Approximation (RA) branch. 
The DA branch constructs a density-driven semantic graph based on the predicted density distribution map. 
It employs GCN to capture feature correlations among regions with similar densities, enhancing the model's robustness to density variations and reducing issues of missed and duplicate counts.
The RA branch generates a representation-driven semantic graph by calculating global representation similarities. Then, the GCN is utilized to explore feature correlations on a global scale, allowing the model to learn diverse individual representations effectively, avoiding the model overfitting to specific patterns and further strengthening model robustness in multi-view and multi-posture scenarios. 
Extensive comparative and ablation experiments on three datasets demonstrate that DSGC-Net outperforms existing methods, achieving state-of-the-art performance.

Overall, the main contributions of this paper are as follows:

1) We propose a novel crowd counting framework called DSGC-Net, which leverages feature correlation mining to improve the accuracy and robustness of crowd counting in complex scenes.

2) We design a dual-stream architecture with a density approximation branch and a representation approximation branch to dynamically adapt to density variations across different regions and capture complex semantic relationships among individuals.

3) Extensive experiments on three widely-used datasets, including ShanghaiTech Part A \cite{zhang2016single}, Part B \cite{zhang2016single}, and UCF-QNRF \cite{idrees2018composition}, demonstrate that the proposed method achieves state-of-the-art performance in crowd counting.

\section{Related Works}

\subsection{Crowd Counting Paradigms}
Mainstream crowd counting methods can be broadly grouped into two types of paradigms: density map-based and point regression-based. Density map-based approaches employ a decoder to predict a continuous crowd density map whose integral yields the estimated counts. In such a paradigm, various methods have been proposed. CSRNet \cite{li2018csrnet} enlarges the receptive field via dilated convolutions and delivers robust estimations in highly congested scenes, while DM-Count \cite{wang2020distribution} replaces hand-tuned Gaussian kernels with an optimal-transport loss, markedly enhancing cross-scenario generalization. However, these methods focus on optimizing the global head count, struggle to deliver point-level localization, and degrade noticeably in sparse or unevenly distributed crowds \cite{Wan_2021_CVPR}. 
To overcome this limitation, point regression-based methods have emerged in recent years. P2PNet \cite{song2021rethinking} reformulates the counting task as a point set prediction problem and uses the Hungarian algorithm to establish a one-to-one correspondence between predicted and annotated points, thereby substantially improving localization accuracy without sacrificing count precision. Building upon this, CAAPN \cite{liu2024consistency} designs a multi-scale anchor pyramid and enforces consistency constraints to adaptively model regions of different densities, further improving robustness in complex scenes. Although point regression-based methods excel at localization, they may still miss targets under heavy occlusion. Motivated by these trade-offs, the proposed DSGC-Net integrates both density map prediction and point-level supervision, using density information to guide point localization for more reliable and precise crowd counting.

\subsection{GNNs for Visual Correlation Modeling}
In recent years, Graph Neural Networks (GNNs) have been preliminarily introduced into the crowd counting task to model long-range dependencies in non-Euclidean spaces. Unlike conventional CNNs, which are restricted to local receptive fields, graph structures can connect distant yet semantically related regions, thus they are capable of efficiently modeling global correlations. TopoCount \cite{abousamra2021localization} introduces a topological consistency constraint that aligns the predicted point graph with the ground-truth topology, effectively reducing merged or duplicate detections. MDGCN \cite{miao2023multi} constructs a multi-scale dynamic graph to strengthen information propagation between sparse and dense regions, yielding significant gains in overall accuracy. HyGNN \cite{luo2020hybrid} further unifies density and localization features as mixed graph nodes and defines both cross-scale and cross-task edges, enabling joint reasoning that simultaneously benefits counting and localization. These aforementioned methods have demonstrated the potential of GNNs in crowd counting. 
However, these methods simply incorporate GNNs into existing counting models as a feature extraction module, without specifically designing them for the characteristics and challenges of the crowd counting task. In our proposed DSGC-Net, starting from the core challenges of the crowd counting task, we model dual graph structures from two perspectives: density distribution and representation approximation, in order to accurately capture the vast change in density and representation distribution within scenes.

\section{Methodology}

\subsection{Overall Framework}
\begin{figure*}[t]
    \centering
    \includegraphics[width=1.0\textwidth]{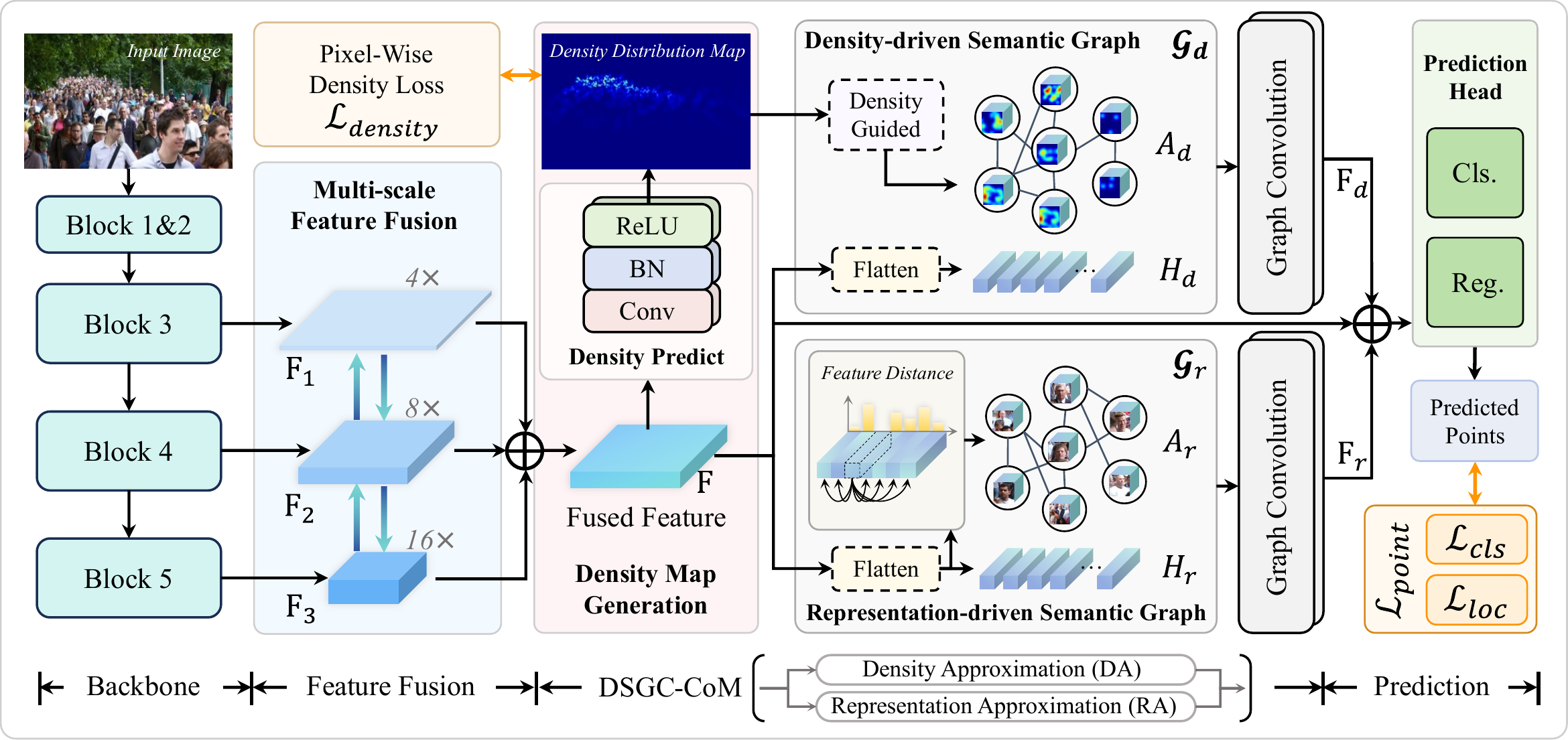} 
    \caption{The overall framework of the proposed Dual-Stream Graph Convolutional Network (DSGC-Net).}
    \label{fig:framework}
\end{figure*}
Fig. \ref{fig:framework} shows the overall framework of the proposed DSGC-Net. It consists of six key steps: 1) The backbone extracts multi-scale features $ \mathbf{F}_1, \mathbf{F}_2, \mathbf{F}_3 $ from the input image. 2) The Path Aggregation Feature Pyramid Network (PA-FPN) \cite{liu2018path} fuses the multi-scale features, integrating high-level semantics and low-level details to generate a fused feature $ \mathbf{F} $. 3) Then, $ \mathbf{F} $ is fed into the density prediction module to generate a density distribution map $ \mathbf{M} $, which provides pixel-wise density information. 4) The \textbf{Dual-Stream Graph Convolutional Correlation Mining (DSGC-CoM)} serves as the core of the framework, encompassing the construction and modeling of dual-stream semantic graphs. Specifically, the Density-driven Semantic Graph (DSG) $ \mathcal{G}_{d} $ is constructed based on the local density similarity across regions, using the previously generated $ \mathbf{M} $, while the Representation-driven Semantic Graph (RSG) $ \mathcal{G}_{r} $ is built based on the global representational similarity of $\mathbf{F}$. 5) Graph Convolutional Networks (GCNs) are applied to $ \mathcal{G}_{d} $ and $ \mathcal{G}_{r} $ separately to capture the latent correlations and generate the enhanced features $ \mathbf{F}_{d} $ and $ \mathbf{F}_{r} $. 6) Finally, $\mathbf{F}_{d}$ and $\mathbf{F}_{r}$ are fused with $\mathbf{F}$, and then fed into the prediction head (regression head and classification head) \cite{song2021rethinking} to localize target points.

\subsection{DSGC-CoM}

The DSGC-CoM consists of two branches: the Density Approximation (DA) and the Representation Approximation (RA). The DA branch constructs the Density-driven Semantic Graph (DSG) guided by predicted pixel-wise density, enabling the aggregation of features from regions with similar densities. Meanwhile, the RA branch constructs the Representation-driven Semantic Graph (RSG) based on representation similarity in the feature space. The following sections introduce the generation of the density distribution map, the construction of the two semantic graphs, and feature correlation mining via GCNs.

\subsubsection{Generation of Density Distribution Map.}
After the input image is processed through the backbone and the PA-FPN, a fused feature $\mathbf{F} \in \mathbb{R}^{C \times H \times W}$ is obtained.
Then, $\mathbf{F}$ is fed into a multi-layer convolutional block to predict the density distribution map $\mathbf{M} \in \mathbb{R}^{1 \times H \times W}$:
\begin{equation}
\mathbf{F}^{(t+1)} 
= 
\operatorname{ReLU}\Bigl(\operatorname{BN}_{t}\bigl(\operatorname{Conv}_{t}\bigl(\mathbf{F}^{(t)}\bigr)\bigr)\Bigr),
\end{equation}
where $\operatorname{Conv}_{t}$ represents a $3 \times 3$ convolutional layer, and $\operatorname{BN}_{t}$ denotes batch normalization. During training, the predicted density distribution map is supervised using a pixel-wise loss function (see Section 3.3).

\subsubsection{Construction of DSG.}
Given the feature map $\mathbf{F} \in \mathbb{R}^{C \times H \times W}$ and the generated density distribution map $\mathbf{M} \in \mathbb{R}^{1 \times H \times W}$, each pixel in $\mathbf{F}$ and $\mathbf{M}$ corresponds to the semantic and density information of a region patch of size $s \times s$ in the original image. Therefore, we can regard each pixel in the feature map as a node in a semantic graph, and the feature vector of each pixel can be regarded as a node feature. In the DA branch, the DSG $ \mathcal{G}_{d} $ is constructed based on the similarity of the density values, making regions with similar densities connected by edges. Specifically, the generated density distribution map $\mathbf{M}$ is first flattened into $\mathbf{M}_{flat} \in \mathbb{R}^{N}$, where $N = H \times W$ (this equation holds throughout the subsequent content). The pairwise density difference between pixels is then computed using the $L_{1}$ norm, resulting in the density similarity matrix:
\begin{equation}
\mathbf{S}_{i,j}
=
\bigl|
\mathbf{M}_{flat,i}-\mathbf{M}_{flat,j}
\bigr|.
\end{equation}
Based on this metric, for each pixel $i$, only the $K$ nearest neighboring pixels are selected, and their corresponding positions in the adjacency matrix are set to 1, while all others are set to 0. The adjacency matrix is thus constructed, and self-loops are added to obtain $\mathbf{A}_{d} \in \mathbb{R}^{N \times N}$. Finally, the DSG is denoted as: $\mathcal{G}_d=\{\mathbf{H}_d, \mathbf{A}_d\}$, where the $\mathbf{H}_d \in \mathbb{R}^{N \times C}$ is the flattened form of $\mathbf{F}$.

\subsubsection{Construction of RSG.}
In the RA branch, the focus is on capturing global semantic similarity in the feature space. By leveraging the high-dimensional representations of pixels in the fused feature map $\mathbf{F} \in \mathbb{R}^{C \times H \times W}$, adjacency relationships are established across spatial locations to connect regions with similar representations, thereby constructing the RSG $ \mathcal{G}_{r} $. Specifically, the fused feature map $\mathbf{F}$ is first flattened to $\mathbf{F}_{flat} \in \mathbb{R}^{N \times C}$, where each row vector represents the representation of a node. The cosine similarity is used to measure the distance between features, yielding the representation distance matrix, formulated as:
\begin{equation}
\mathbf{Dist}_{i,j}
=
\frac{
\langle
\mathbf{F}_{flat,i},
\mathbf{F}_{flat,j}
\rangle
}{
\|
\mathbf{F}_{flat,i}
\|
\,
\|
\mathbf{F}_{flat,j}
\|
},
\end{equation}
where $\langle \cdot, \cdot \rangle$ denotes the inner product, and $||\cdot||$ represents the $L_{2}$ norm. The $K$ most similar neighboring nodes are then selected based on this similarity measure, forming a sparse adjacency matrix as in the previous method. Self-loops are subsequently added to obtain $\mathbf{A}_{r}\in \mathbb{R}^{N \times N}$, thereby establishing connections between regions with similar representations. After the above process, the constructed RSG is denoted as: $\mathcal{G}_r=\{\mathbf{H}_r, \mathbf{A}_r\}$, where the $\mathbf{H}_r \in \mathbb{R}^{N \times C}$ is the flattened form of $\mathbf{F}$.

\subsubsection{Feature Correlation Mining via GCNs.}
For both the DA branch and the RA branch, once the adjacency matrix $\mathbf{A}$ is obtained ($\mathbf{A}_d^{(0)}$ for DSG, and  $\mathbf{A}_r^{(0)}$ for RSG), it is fed into GCNs along with the corresponding node feature matrix $\mathbf{H}^{(0)} \in \mathbb{R}^{N \times C}$ ($\mathbf{H}_d^{(0)}$ for DSG, and  $\mathbf{H}_r^{(0)}$ for RSG). The GCNs progressively aggregate the features of neighboring nodes within the graph structure defined by the adjacency matrix, thereby capturing richer contextual semantics. It can be formulated as follows:
\begin{equation}
\mathbf{H}^{(l+1)}
=
\sigma
\Bigl(
\mathbf{D}^{-\frac{1}{2}}\,
\mathbf{A}\,
\mathbf{D}^{-\frac{1}{2}}\,
\mathbf{H}^{(l)}
\,
\mathbf{W}^{(l)}\Bigl),
\end{equation}
where $\mathbf{D}_{ii} = \sum_{j} \mathbf{A}_{ij}$, $\mathbf{W}^{(l)}$ are learnable parameters, and the activation function $\sigma(\cdot)$ is chosen as $\operatorname{ReLU}$. Note that the GCNs in the two branches have independent weights. After multiple propagation layers, the resulting $\mathbf{H}^{(l)}$ integrates contextual features from its neighborhood and is reshaped into the form of $ \mathbb{R}^{C_{out} \times H \times W} $, yielding the final graph convolutional features $\mathbf{F}_{d}$ and $\mathbf{F}_{r}$.

\subsection{Joint Loss Function}

We design a joint loss function that combines density supervision and point supervision to optimize both tasks simultaneously. The density loss $\mathcal{L}_{density}$ is measured using Mean Squared Error (MSE) between the predicted density distribution map and the ground truth density distribution map, ensuring the model's accuracy in pixel-wise local density estimation. Formally, it is defined as:
\begin{equation}
\mathcal{L}_{density} = \frac{1}{N} \sum_{i=1}^N \| \mathbf{M}_i - \mathbf{M}_i^{gt} \|^2.
\end{equation}
Meanwhile, the point supervision loss $\mathcal{L}_{point}$, as referenced in \cite{song2021rethinking}, consists of two components. The loss in the regression head $\mathcal{L}_{loc}$ employs Euclidean Loss to constrain the regression accuracy of target points. In contrast, the loss in the classification head $\mathcal{L}_{cls}$ adopts Cross-Entropy Loss to supervise the classification of target points. It is defined as follows:
\begin{align}
    \mathcal{L}_{point} &= \mathcal{L}_{cls} + \lambda_1 \mathcal{L}_{loc} \notag \\
    &= - \frac{1}{M} 
    \left( 
        \sum_{i=1}^N \log \hat{c}_{\xi(i)}  
        + \lambda_2 \sum_{i=N+1}^M \log \left( 1 - \hat{c}_{\xi(i)} \right) 
    \right) \notag \\
    &\quad + \frac{\lambda_1}{N} \sum_{i=1}^N \| p_i - \hat{p}_{\xi(i)} \|^2,
\end{align}
where $\lambda_1$ is a weight term for balancing the effect of the location regression loss, and $\lambda_2$ is a reweight factor for negative proposals in the classification. $||\cdot||$ denotes the $L_{2}$ norm. Finally, the joint loss function for the entire model's optimization is: $\mathcal{L}_{joint} = \mathcal{L}_{point} +  \mathcal{L}_{density}$.
\section{Experiments}
\subsection{Experimental Setup}
To verify the effectiveness of the proposed method, we conduct experiments on three of the most widely used datasets: ShanghaiTech Part A \cite{zhang2016single} and Part B
\cite{zhang2016single} and UCF-QNRF \cite{idrees2018composition}, each exhibiting different levels of scene complexity and density distribution characteristics. 
Following previous works \cite{liu2024consistency,zhang2016single,lei2024ddranet}, we adopt Mean Absolute Error (MAE) and Mean Squared Error (MSE) as evaluation metrics.

We use the VGG-16 \cite{simonyan2015very} with pre-trained weights as the backbone network by default. In the experiments, data augmentation includes random scaling, random cropping, and random contrast disturbance. During training, we set the batch size to 8. In the loss function, $\lambda_1$ and $\lambda_2 $ are adopted from \cite{song2021rethinking} and set to $2 \times 10^{-4}$ and 0.5, respectively. The learning rate is set to $1 \times 10^{-4}$, and the model parameters are updated using the Adam optimization algorithm \cite{kingma2014adam}. Since the VGG-16 backbone is pre-trained on ImageNet \cite{deng2009imagenet}, its learning rate is set to $1 \times 10^{-5}$. By default, the number of selected neighboring nodes $K$ in semantic graph construction is set to 4. Note that due to the extremely large image size of the UCF-QNRF dataset, we constrain the longer edge to a maximum of 1,408 pixels while maintaining the aspect ratio. 

\begin{table}[t]
    \centering
    \caption{Comparison of counting performance with state-of-the-art methods. The best and second-best results are highlighted in \textcolor{red}{red} and \textcolor{blue}{blue}, respectively.}
    \label{tab:comparison}
    \renewcommand{\arraystretch}{1.15}
    \small
    \setlength{\tabcolsep}{0.35mm}
    \rowcolors{1}{gray!15}{white}
    \begin{tabular}{l c c c c c c c c}
        \hiderowcolors
        \toprule
            \noalign{\vskip -2.5pt}
        \multirow{2}{*}{Method} & 
        \multirow{2}{*}{Venue} & 
        \multirow{2}{*}{Backbone} & 
        \multicolumn{2}{c}{SHTechPartA} &
        \multicolumn{2}{c}{SHTechPartB} &
        \multicolumn{2}{c}{UCF-QNRF} \\
        \noalign{\vskip -2.5pt}
        \cmidrule(lr){4-5}\cmidrule(lr){6-7}\cmidrule(lr){8-9}
        \noalign{\vskip -2.5pt}
         &  &  & MAE & MSE & MAE & MSE & MAE & MSE \\
        \noalign{\vskip -2.5pt}
        \midrule
        \noalign{\vskip -3pt}
        \showrowcolors
        CSRNet\cite{li2018csrnet} & CVPR'18 & VGG-16\cite{simonyan2015very} & 68.2 & 115.0 & 10.6 & 16.0 & - & - \\
        
        BL\cite{ma2019bayesian} & ICCV'19 & VGG-19\cite{simonyan2015very} & 62.8 & 101.8 & 7.7 & 12.7 & 88.7 & 154.8 \\
        
        DM-Count\cite{wang2020distribution} & NIPS'20 & VGG-19\cite{simonyan2015very} & 59.7 & 95.7 & 7.4 & 11.8 & 85.6 & 148.3 \\
        HYGNN\cite{luo2020hybrid} & AAAI'20 &VGG-16\cite{simonyan2015very} & 60.2 & 94.5 & 7.5 & 12.7 & 100.8 & 185.3 \\
        P2PNet\cite{song2021rethinking} & ICCV'21 & VGG-16\cite{simonyan2015very} & 52.7 & 85.1 & 6.2 & 9.9 & 85.3 & 154.5 \\
        TopoCount\cite{abousamra2021localization}     & AAAI'21   & VGG-16\cite{simonyan2015very}      & 61.2 & 104.6 & 7.8  & 13.7 & 89.0    & 159.0   \\
        LSC-CNN\cite{sam2021locate} & PAMI'21 & VGG-16\cite{simonyan2015very} & 66.4 & 117.0 & 8.1 & 12.7 & 120.5 & 218.2 \\
        CLTR\cite{liang2022end} & ECCV'22 & DETR\cite{carion2020end} & 56.9 & 95.2 & 6.5 & 10.6 & 85.8 & 141.3 \\
        Ctrans-MISN\cite{zeng2022joint} & PRAI'22 & ViT\cite{dosovitskiy2020image} & 55.8 & 95.9 & 7.3 & 11.4 & 95.2 & 180.1 \\
        NDConv\cite{zhong2022improved} & SPL'22 & ResNet-50\cite{He_2016_CVPR} & 61.4 & 104.2 & 7.8 & 13.8 & 91.2 & 165.6 \\
        AutoScale\cite{xu2022autoscale} & IJCV'22 & VGG-16\cite{simonyan2015very} & 65.8 & 112.1 & 8.6 & 13.9 & 104.4 & 174.2 \\
        
        PTCNet\cite{liu2023multi}       & EAAI'23   & PVT\cite{wang2021pyramid} & \textcolor{blue}{51.7} & 79.6 & 6.3  & 10.6 & \textcolor{blue}{79.7} & \textcolor{red}{133.2} \\
        GMS\cite{wang2023crowd} & TIP'23 & HRNet\cite{wang2021deep} & 68.8 & 138.6 & 16.0 & 33.5 & 104.0 & 197.4 \\
        DMCNet\cite{wang2023dynamic} & WACV'23 & VGG-16\cite{simonyan2015very} & 58.5 & 84.6 & 8.6 & 13.7 & 96.5 & 164.0 \\
        
        VMambaCC\cite{ma2024vmambacc} & arXiv'24 & Mamba\cite{gu2024mamba} & 51.9 & 81.3 & 7.5 & 12.5 & 88.4 & 144.7 \\
        DDRANet\cite{lei2024ddranet} & SPL'24 & VGG-16\cite{simonyan2015very} & 52.1 & \textcolor{blue}{78.4} & 6.9 & 10.3 & 89.2 & 146.9 \\
        CAAPN\cite{liu2024consistency}        & PAMI'24  & VGG-16\cite{simonyan2015very}      & 54.4 & 97.3  & \textcolor{red}{5.8} & \textcolor{blue}{9.8} & 83.9  & 144.3 \\
        CrowdFPN\cite{yu2025crowdfpn} & Ap.Int.'25 & Twins\cite{NEURIPS2021_4e0928de} & 52.5 & 88.5 & 6.5 & 9.9 & 81.2 & 157.3 \\
        CC-SoftHGNN\cite{lei2025softhgnn} & arXiv'25 & PVT\cite{wang2021pyramid} & \textcolor{blue}{51.7} & 79.2 & 6.6 & 10.5 & - & - \\
        \noalign{\vskip -2pt}
        \midrule
        \noalign{\vskip -3pt}
        \textbf{DSGC-Net} & \textbf{PRCV'25} & \textbf{VGG-16}\cite{simonyan2015very} & \textbf{\textcolor{red}{48.9}} & \textbf{\textcolor{red}{77.8}} & \textbf{\textcolor{blue}{5.9}} & \textbf{\textcolor{red}{9.3}} & \textcolor{red}{\textbf{79.3}} & \textbf{\textcolor{blue}{133.9}} \\
        \noalign{\vskip -2pt}
        \bottomrule
        \noalign{\vskip -2pt}
    \end{tabular}
\end{table}

\subsection{Comparison with Other Methods}
We compare DSGC-Net with various representative methods, as shown in Table \ref{tab:comparison}. On ShanghaiTech Part A, our method achieves the best performance, with MAE and MSE of 48.9 and 77.8, respectively. Notably, the MAE is improved by 5.4\% compared to the second-best method. Furthermore, for sparse scenes such as Part B, our method achieves the second-best MAE of 5.9 and the best MSE of 9.3, with MSE improving by 5.1\% over the second-best method. On the UCF-QNRF dataset, our method achieves the best MAE of 79.3 and the second-best MSE of 133.9.
Overall, our approach maintains significant advantages across all datasets.

\subsection{Ablation Study}
\begin{table}[t]
    \centering
    \setlength{\tabcolsep}{3mm}
    \caption{Ablation experiment results on the effectiveness of each component.}
    \label{tab:ablation_modules}
    \fontsize{9pt}{11pt}\selectfont
    \begin{tabular}{l c c c c c}
        \toprule
        \noalign{\vskip -1pt}
        \multirow{2}{*}{Module} & 
        \multicolumn{2}{c}{ShanghaiTech Part A} & 
        \multicolumn{2}{c}{ShanghaiTech Part B} &
        \multirow{2}{*}{Param.(M)} \\
        \noalign{\vskip -2pt}
        \cmidrule(r){2-3}\cmidrule(r){4-5}
        \noalign{\vskip -1pt}
         & MAE & MSE & MAE & MSE & \\
        \noalign{\vskip -2pt}
        \midrule
        \noalign{\vskip -1pt}
        Baseline     & 53.6 & 85.1 & 7.4 & 11.6 & 22.3M \\
        Add DP       & 51.3 & 79.9 & 6.6 & 10.7 & +2.3M \\
        Add DP \& DA & 49.8 & 79.5 & 6.5 & 10.2 & +2.6M \\
        Add RA       & 52.0 & 83.7 & 6.9 & 11.1 & +0.3M \\
        \textbf{Add All}      & \textbf{48.9} & \textbf{77.8} & \textbf{5.9} & \textbf{9.3} & \textbf{+2.9M} \\
        \noalign{\vskip -1pt}
        \bottomrule
    \end{tabular}
\end{table}

\subsubsection{Effectiveness of Each Component.}
\begin{figure}[t]  
    \centering
    \includegraphics[width=1.0\linewidth]{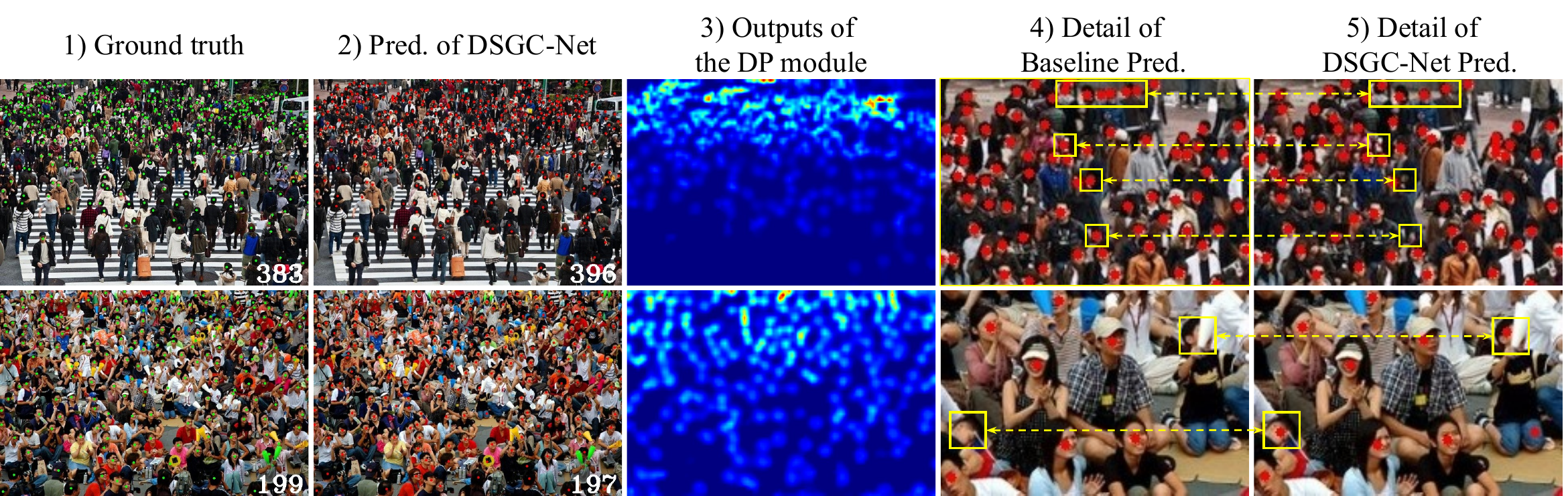}
    \caption{Visualization examples. 1) The ground truth. 2) Prediction results of the proposed DSGC-Net. 3) The outputs of the Density Prediction (DP) module. 4) and 5) are the detailed comparisons of the prediction results of the baseline and the proposed DSGC-Net. }
    \label{fig:vis}
\end{figure}

To verify the effectiveness of each component, we conduct five groups of ablation experiments on the ShanghaiTech Part A and Part B datasets, as shown in Table \ref{tab:ablation_modules}. 1) Baseline: removing all the proposed modules. 2) Add only the Density Prediction (DP) module. 3) Add both the DP module and the DA branch. 4) Add only the RA branch. 5) Full DSGC-Net model. The experimental results demonstrate that the proposed modules improve the model performance significantly, especially in dense scenes (Part A).
It is worth noting that the proposed DA and RA branches have only 0.3M parameters respectively, indicating that our approach shows excellent performance while maintaining lightweight.
Moreover, Fig. \ref{fig:vis} shows visualization results from two examples.
In example 1, our method avoids duplicate counts in high-density regions, while in example 2, it reduces the risk of missed counts under complex conditions such as occlusion and diverse postures.

\begin{figure}[t]
    \centering
    \begin{subfigure}{0.48\linewidth}
        \centering
        \includegraphics[width=\linewidth]{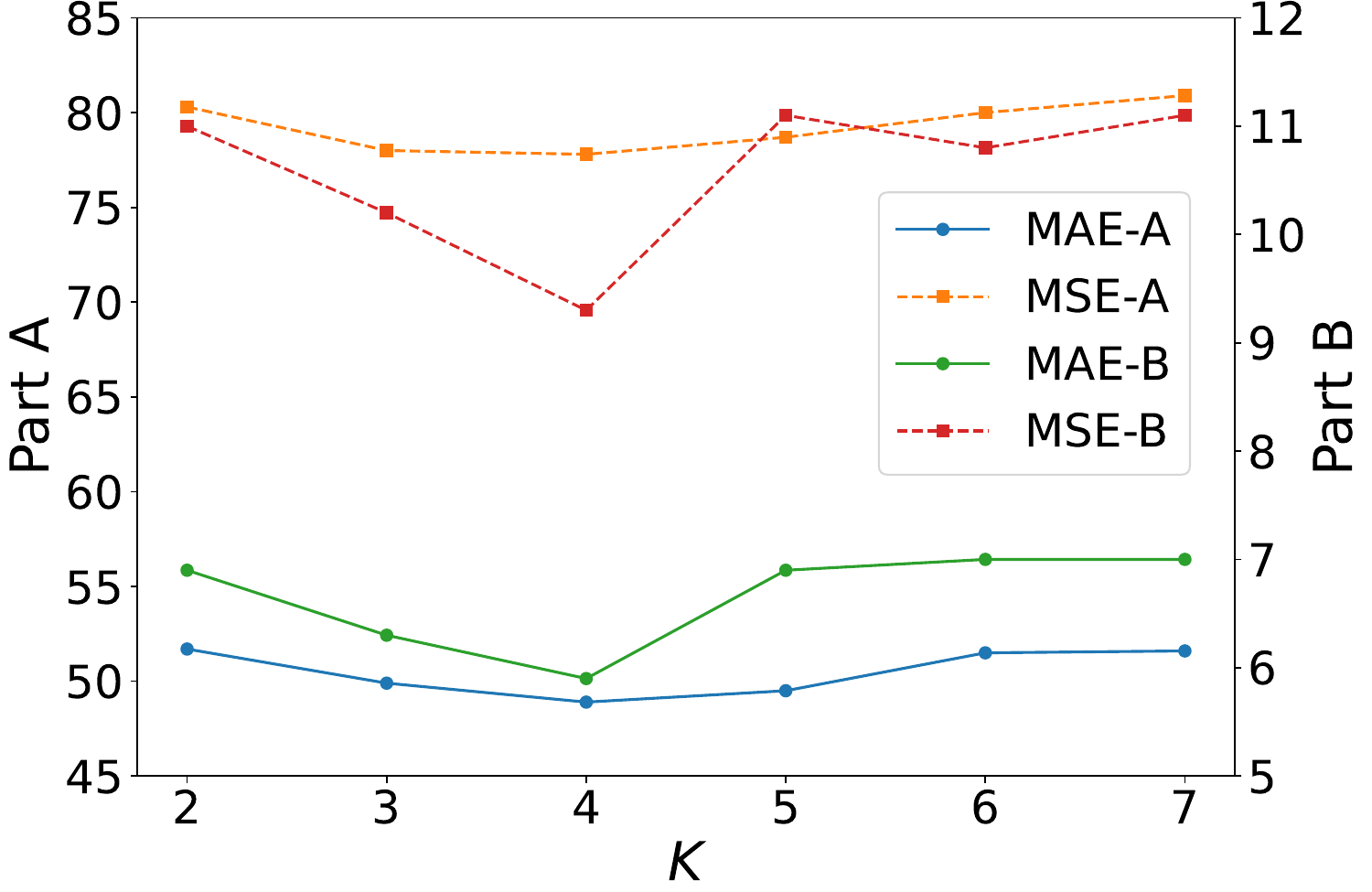}
        \caption{Impact of the number of neighboring nodes ($K$) on performance on the ShanghaiTech Part A and Part B datasets.}
        \label{fig:k_choice}
    \end{subfigure}
    \hfill
    \begin{subfigure}{0.48\linewidth}
        \centering
        \includegraphics[width=\linewidth]{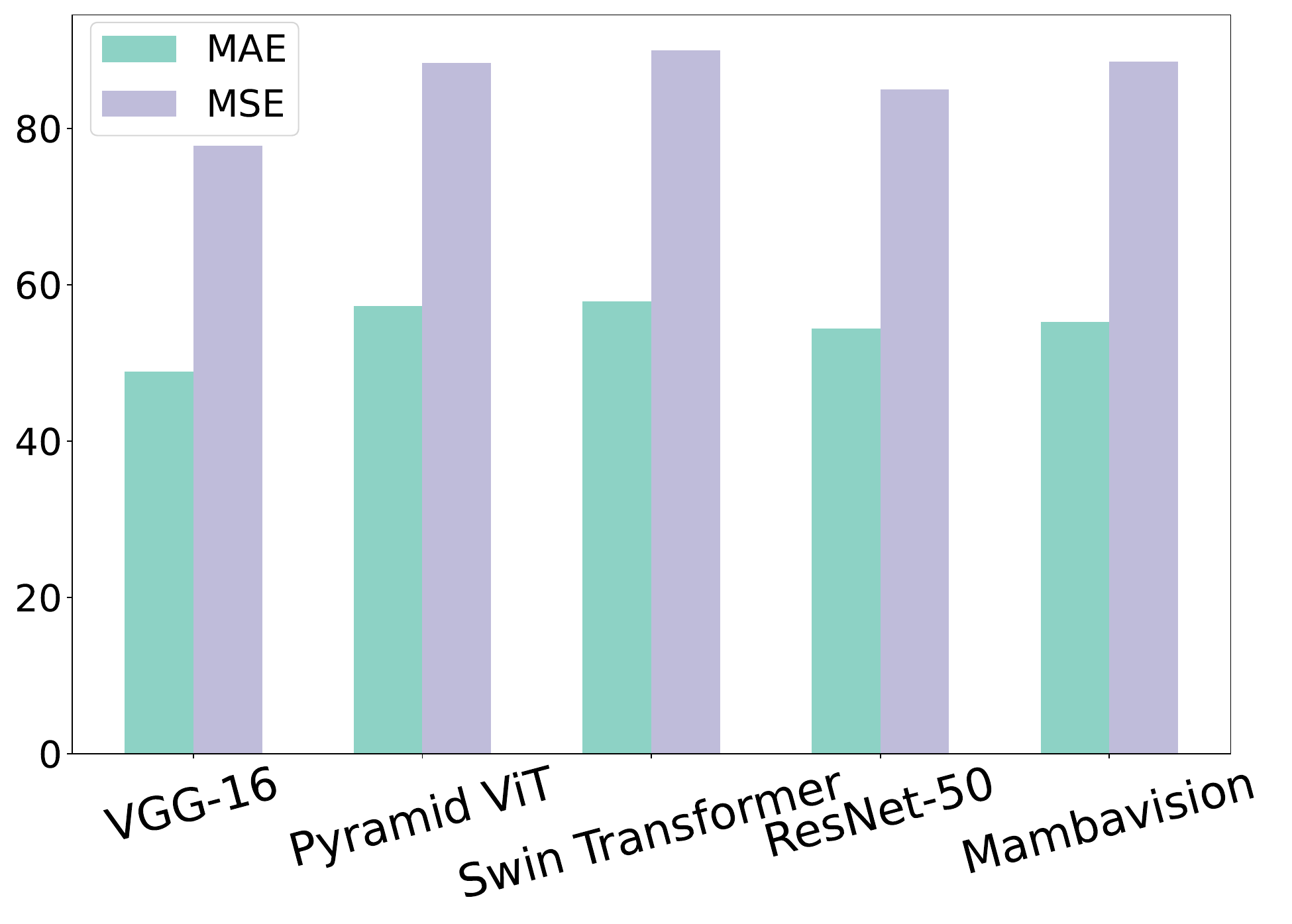}
        \caption{Impact of different backbones on the ShanghaiTech Part A dataset.}
        \label{fig:backbone_choice}
    \end{subfigure}
    \caption{Ablation experiments on (a) the number of neighboring nodes, and (b) using different backbones.}
    \label{fig:both}
\end{figure}
\subsubsection{Selection of the Number of Neighboring Nodes $K$.}
Figure. \ref{fig:k_choice} illustrates how varying the number of neighbouring nodes $K$ affects performance. Accuracy peaks at $K = 4$ on both datasets, with noticeable fluctuations elsewhere. A small $K$ renders the graph overly sparse, hindering the capture of long-range correlations, whereas an excessively large $K$ introduces many semantically irrelevant edges, injecting noise into feature aggregation. Thus, a moderate neighbourhood size strikes the best balance between information coverage and robustness.

\subsubsection{Using Different Backbones.}
To evaluate the impact of backbone selection on model performance, we experimented with five widely used feature extraction architectures: VGG-16 \cite{simonyan2015very}, Pyramid ViT \cite{wang2021pyramid}, Swin Transformer \cite{Liu_2021_ICCV}, ResNet-50 \cite{He_2016_CVPR}, and Mambavision \cite{Hatamizadeh_2025_CVPR}, conducting experiments on the ShanghaiTech Part A dataset, as shown in Figure. \ref{fig:backbone_choice}. The results demonstrate that our framework exhibits strong generalizability, effectively adapting to various feature extraction networks while achieving significant performance improvements. Notably, choosing VGG-16 delivers the best counting accuracy. 
Moreover, experiments with Mambavision and other emerging backbones provide valuable insights for future research.

\begin{table}[t]
    \centering
    \setlength{\tabcolsep}{3.5mm}
    \caption{Impact of using different graph operators.}
    \label{tab:gragh_methods}
    \fontsize{9pt}{11pt}\selectfont
    \begin{tabular}{l c c c c c}
        \toprule
        \noalign{\vskip -1pt}
        \multirow{2}{*}{GNNs} & 
        \multicolumn{2}{c}{ShanghaiTech Part A} & 
        \multicolumn{2}{c}{ShanghaiTech Part B} &
        \multirow{2}{*}{FLOPs(G)} \\
        \noalign{\vskip -2pt}
        \cmidrule(r){2-3}\cmidrule(r){4-5}
        \noalign{\vskip -1pt}
         & MAE & MSE & MAE & MSE & \\
        \noalign{\vskip -2pt}
        \midrule
        \noalign{\vskip -1pt}
        GAT \cite{velickovic2017graph}     & 50.6 & 81.2 & 6.6 & 10.3 & 15.1\\
        HGNN \cite{gao2023hgnn}    & 50.9 & 77.7 & 6.8 & 10.9 & 15.5 \\
        \textbf{GCN} \cite{kipf2016semi}     & \textbf{48.9} & \textbf{77.8} & \textbf{5.9} & \textbf{9.3} & \textbf{15.1}\\
        \noalign{\vskip -1pt}
        \bottomrule
    \end{tabular}
\end{table}
\subsubsection{Using Different Graph Operators.}

Within the dual-stream architecture, we successively replaced the graph message-passing module with GCN \cite{kipf2016semi}, GAT \cite{velickovic2017graph} , and HGNN \cite{gao2023hgnn} with higher-order interaction modeling capability, while uniformly setting the number of neighboring nodes to 4. The performance of each variant is summarized in Table \ref{tab:gragh_methods}, with FLOPs measured under an input resolution of $3\times128\times128$. Experimental results demonstrate that all three operators deliver strong performance, demonstrating the robustness of DSGC-Net and its general utility as a feature enhancement framework for crowd counting. Notably, GCN delivers the best trade-off between accuracy and efficiency, attributed to its stable aggregation mechanism that enhances feature representation while effectively suppressing noise and mitigating over-smoothing.

\section{Conclusion}

In this paper, we propose a Dual-Stream Graph Convolutional Network (DSGC-Net) for crowd counting, which is a hybrid architecture that integrates both density map-based and point regression-based strategies. DSGC-Net enhances the ability to perceive local density variations through the density approximation branch, while the representation approximation branch effectively mitigates overfitting to specific representation types, thereby improving adaptability to various individual representations. Experimental results demonstrate that our method significantly enhances counting accuracy and model robustness, providing a new insight and technical solution for crowd counting tasks.

\subsubsection{Acknowledgements:} This work was supported by the National Natural Science Foundation of China (62403345), and the Shanxi Provincial Department of Science and Technology Basic Research Project (202403021212174).

\bibliographystyle{splncs04}
\bibliography{a_cite}
\end{document}